\documentclass{svproc}
\usepackage[utf8]{inputenc}
\usepackage{graphicx}
\usepackage{makecell}
\usepackage{nth}
\usepackage{hyperref}
\usepackage[table]{xcolor}
\usepackage{longtable} 
\usepackage{booktabs}
\urldef{\mails}\path|felber@student.tugraz.at|

\newcommand{\etal}{\textit{et al.}}

\begin{document}
%
\title{Constraint 2021: Machine Learning Models for COVID-19 Fake News Detection Shared Task}
\titlerunning{Machine Learning Models for COVID-19 Fake News Detection}
%
%
\author{Thomas Felber}
%
%

\institute{
Institute of Interactive Systems and Data Science \\
Graz University of Technology \\
Inffeldgasse 13, 8010 Graz, Austria\\
\mails
}

\maketitle              
\begin{abstract}
In this system paper we present our contribution to the
Constraint 2021 COVID-19 Fake News Detection Shared Task, which poses
the challenge of classifying COVID-19 related social media posts as either \textit{fake} or 
\textit{real}. In our system, we address this challenge by applying
classical machine learning algorithms together with
several linguistic features, such as n-grams, readability, emotional tone and punctuation.
In terms of pre-processing, we experiment with various steps like
stop word removal, stemming/lemmatization, link removal and more.
We find our best performing system to be based on a linear SVM, which obtains a
weighted average F1 score of 95.19\% on test data, which lands a place
in the middle of the leaderboard (place 80 of 167).

\keywords{Machine Learning, Classification, Supervised Learning}
\end{abstract}
\section{Introduction}
The internet functions as a valuable resource for  
individuals who seek information online.
A majority of people for instance use it when searching for health-related information 
\cite{doi:10.1177/0898264311428167}. With the rise of social media in recent years,
platforms like Facebook, Instagram and Twitter 
are a key resource to go
for following updates during crisis events \cite{palen2008online}
like the COVID-19 pandemic spreading. On these platforms
any registered user can publish and disseminate unverified content. This can act as 
a breeding ground for disinformation and fake news, potentially exposing millions of users
to harmful misinformation in a short frame of time \cite{marwick2017media}.
In order to prevent the negative impact of misinformation spread, there is an incentive 
to develop new methods capable of assessing the validity of social media posts.

Constraint 2021 \footnote{\url{http://lcs2.iiitd.edu.in/CONSTRAINT-2021/}}
is a workshop on combating online hostile posts (hate speech, fake news detection, cyberbullying, etc.) in regional languages
during emergency situation and is held as part of the \nth{35} AAAI conference on artificial intelligence
on February 8, 2021. It encourages researchers from interdisciplinary domains to think of new ways on
how to combat online hostile posts  
during emergencies such as the 2021 US presidential election or the COVID-19 pandemic spreading.
In alignment with that, the workshop organizes a shared 
task\footnote{\url{https://constraint-shared-task-2021.github.io/}} \cite{patwa2021overview} and invites for participation in it's 
two subtasks, which focus on COVID-19 Fake News Detection in English (Subtask 1) and 
Hostile Post Detection in Hindi (Subtask 2).

In this paper, we discuss our take on Subtask 1, where we apply classical machine learning
algorithms with diverse groups of linguistic features, considering n-grams, 
readability, emotional tone and punctuation, in order to classify social media posts
as either \textit{fake} or \textit{real}. 
\section{Related Work}
Interest about fake news detection has increased over the last years,
and there are various shared tasks in this domain, like 
Profiling Fake News Spreaders on Twitter \cite{rangel2020overview},
Fake News Challenge (FNC-1) \cite{pomerleau2017fake},
RumorEval \cite{derczynski2017semeval,gorrell2019semeval},
CheckThat! \cite{barron2018overview},
and Fact Extraction and Verification (FEVER) \cite{thorne2018fact,thorne2019second}. 

In Profiling Fake News Spreaders on Twitter, the task aims at identifying possible fake
news spreaders on social media as a first step towards preventing fake news from 
being propagated among online users.
The organizers of FNC-1 propose their task to be tackled from a stance perspective,
i.e. estimate the stance of a body text from a news article relative to a headline,
where the body text may agree, disagree, discuss or be unrelated to the headline.
This is similar to task-A of RumorEval whose objective is to track
how other sources orient to the accuracy of a rumourous story. In both task-B of RumorEval and
task-B of CheckThat! the goal is to predict the veracity of a given rumor or claim. 
Task-A of CheckThat! is concerned with the detection of check-worthy tweets, i.e. tweets
that include a claim that is of interest to a large audience (specially journalists),
might have a harmful effect, etc. 
The idea behind FEVER is to evaluate the ability of systems to verify factual claims
by extracting evidence from Wikipedia. Based on the extracted evidence, the claims are labeled
Supported, Refuted, or NotEnoughInfo (if there isn’t sufficient evidence to either support 
of refute it).

Through these shared tasks, numerous methods for fake news detection have been found. 
Zhou \etal \cite{zhou2019gear} propose GEAR, 
a graph-based evidence aggregating and 
reasoning framework, that enables information to transfer on a fully-connected evidence graph 
and then utilizes different aggregators to collect multi-evidence information. In conjunction
with BERT \cite{devlin2018bert}, an effective pre-trained language representation model,
they are able to achieve a promising FEVER score of 67.10\%. Umer \etal \cite{umer2020fake} 
address
the problem of fake news stance detection in FNC-1 by following a hybrid neural network 
approach, that combines the capabilities of CNN and LSTM, in addition with two different 
dimensionality reduction approaches, Principle Component Analysis (PCA) and Chi-Square. 
Within their experimental results they show that PCA outperforms Chi-Square and obtain
state-of-the-art performance with 97.8\% accuracy. Ghanem \etal \cite{ghanem2019upv} 
use classical machine learning algorithms together with
stylistic, lexical, emotional, sentiment, meta-structural and Twitter-based features
for the task of rumor stance classification in RumorEval, and introduce two novel 
feature sets that proved to be successful for this task.
\section{Task Description}
The Constraint 2021 Shared Task is composed of two individual subtasks:
\begin{itemize}
    \item\textbf{Subtask 1:} COVID-19 Fake News Detection in English
    \item\textbf{Subtask 2:} Hostile Post Detection in Hindi
\end{itemize}
Since our contribution to the shared task is solely concerned with Subtask 1, we shall now
give a closer definition of that task, which is as follows: 
Given a collection of COVID-19-related posts from various social media platforms such as 
Twitter, Facebook, Instagram, etc., classify the posts as either \textit{fake} or 
\textit{real}, as illustrated in Table \ref{tbl:example}
\rowcolors{2}{green!15}{red!15}
\begin{table}[h!]
    \centering
    \caption{Example classification of social media posts}
    \label{tbl:example}
    \begin{tabular}{*2c}
    \toprule
    Post & Class \\
    \midrule
    If you take Crocin thrice a day you are safe. & fake \\
    Wearing mask can protect you from the virus.  & real \\
    \bottomrule
    \end{tabular}
\end{table}
\\
The basis of the social media posts is provided by a manually annotated
COVID-19 fake news dataset that is described in the following section.
\subsection{Dataset}
The foundation of social media posts provided for
Subtask 1 is laid by 
a fake news dataset \cite{patwa2020fighting} containing 10700 manually
annotated social media posts and articles of real and fake
news on COVID-19.
The dataset is split into training, validation and
test sets with respect to a 3:1:1 ratio as illustrated in Table \ref{tbl:dataset}.
\rowcolors{2}{green!0}{red!0}
\begin{table}[h!]
    \centering
    \caption{Dataset splits and distribution of class labels}
    \label{tbl:dataset}
    \begin{tabular}{ *{4}{@{\hskip8pt}c@{\hskip8pt}} }
    \toprule
    \textbf{Split} & \textbf{Real} & \textbf{Fake} & \textbf{Total} \\
    \midrule
    Training set    & 3360 & 3060 & 6420 \\ \midrule
    Validation set  & 1120 & 1020 & 2140 \\ \midrule
    Test set        & 1120 & 1020 & 2140 \\ \midrule
    \textbf{Total}  & 5600 & 5100 & 10700 \\ 
    \bottomrule
    \end{tabular}
\end{table}
\\
Regarding class label distribution, a consistent balance across training, validation and
test set has been established, where 52.34\% of the samples are attributed to real news
and the remaining 47.66\% to fake news.

Further dataset metrics can be derived by determining total unique words, average words per 
post and average characters per post as shown in Table \ref{tbl:dataset2}.
\begin{table}[h!]
    \centering
    \caption{Further numeric features of the dataset}
    \label{tbl:dataset2}
    \begin{tabular}{ *{4}{@{\hskip8pt}c@{\hskip8pt}} }
    \toprule
    \textbf{Attribute} & \textbf{Fake} & \textbf{Real} & \textbf{Combined} \\
    \midrule
    Unique words        & 19728 & 22916 & 37503 \\ \midrule
    Avg. words per post & 21.65 & 31.97 & 27.05 \\ \midrule
    Avg. chars per post & 143.26 & 218.37 & 182.57 \\ 
    \bottomrule
    \end{tabular}
\end{table}
\\
Based on data from the table, it can be observed that on average, real news posts are
longer than fake news post in terms of both characters per post and words per post.
The size of the vocabulary, given by the unique words in the dataset, is 37505 where 5141
of these words are contained in both fake and real posts.
\subsection{Evaluation}
Each team participating at Subtask 1 has access to training, validation and test splits of the dataset,
where training and validation data are labeled, and test data are unlabeled.
Participant systems should predict class labels for social media posts contained in test data.
Performance is measured in terms of weighted average F1 score, however, weighted average precision,
weighted average recall and accuracy are also reported. The best performing baseline result
provided by the organizers is a linear SVM with 93.46\% weighted average F1 score.
\section{System Description}
In our approach, we experiment with different pre-processing pipelines and
apply classical machine learning algorithms with diverse groups of linguistic features,
such as n-grams, readability, emotional tone and punctuation.
Regarding machine learning, we utilize scikit-learn \cite{pedregosa2011scikit}, 
a Python module providing a wide range of state-of-the-art machine learning algorithms for
medium-scale supervised and unsupervised problems. For some of our pre-processing steps
we utilize functionality provided by the Natural Language Toolkit (NLTK) \cite{loper2002nltk}.
Furthermore, to extract some of the linguistic features, 
we use Linguistic Inquiry and Word Count (LIWC) \cite{pennebaker2015development}, which is
a text analysis program, capable of reading a given text and counting the percentage of words
that reflect different emotions, thinking styles, social concerns, and even parts of speech.
\subsection{Pre-processing}
Using scikit-learn's pipeline capability, we create and experiment with 
different pre-processing pipelines, where in each of these, the following steps
may be applied:
\begin{description}
    \item[Stop word removal:] If this step is applied, English stop words are removed from 
    the social media posts. In that case, the stop word dictionary is provided by NLTK.
    \item[Link removal:] In this step, hypertext links are removed from 
    the social media posts. This is accomplished via regular expressions.
    \item[Lemmatization or stemming:] During this step either lemmatization or stemming is performed.
    Lemmatization is achieved by NLTK's WordNet Lemmatizer and stemming by NLTK's Snowball Stemmer 
    implementation, which is based on Porter2 stemming algorithm \cite{porter2002english}.
    \item[Reply removal:] In this step, words starting with @ (mostly used for Twitter replies) are removed.
    This is also accomplished via regular expressions.
\end{description}
Apart from the pre-processing steps described above, 
the following steps are always applied to every social media post:
\begin{description}
    \item[Lowercase transformation:] To account for differences in capitalization, in this step,
    every word is transformed to lowercase.
    \item[XML entity replacement:] Some social media posts contain XML entities such as 
    "\&amp;", "\&gt;" and "\&lt;". In this step, those entities are replaced by 
    their corresponding text symbol.
\end{description}
\subsection{Features}
In terms of features, we focus on linguistic aspects such as
n-grams, readability, emotional tone and punctuation. Based on these features,
together with scikit-learn's pipeline capability, we create and experiment with different feature
extraction pipelines, where in each of these, the features mentioned above may be contained:
\begin{description}
    \item[N-grams:] We extract unigrams and bigrams derived
    from the bag of words representation of each
    pre-processed social media post. To address differences in 
    content length, we encode these features as TF-IDF values.
    To accomplish this, we make use of scikit-learn's CountVectorizer in conjunction with
    it's TfidfTransformer. Regarding tokenization, we experiment with NLTK's TweetTokenizer
    and WordPunctTokenizer.
    \item[Readability:] Here we extract features that reflect text understandability.
    These features are based on properties such as words per sentence, characters per word, 
    syllables per word, number of complex words, and so on. Various readability metrics
    that calculate a readability score using this information already exist.
    In our experiments, we calculate several of these metrics, including Automatic 
    Readability Index (ARI) \cite{senter1967automated},
    Flesch-Kincaid \cite{flesch2007flesch}, Coleman-Liau \cite{coleman1975computer}, 
    Flesch Reading Ease \cite{flesch1948new} and more.
    \item[Emotional tone:] We use the Linguistic Inquiry and
    Word Count text analysis software (LIWC, Version 1.6.0 2019) \cite{pennebaker2015development} 
    to extract the proportions of words that fall
    into certain psycholinguistic categories 
    (e.g., positive emotions, negative emotions, cognitive processes, social words).
    In our work, we experiment with features derived from LIWC summary categories (e.g. emotional tone, 
    analytical thinking) and psychological categories (e.g. social words, cognitive processes).
    \item[Punctuation:] Also using LIWC, we construct a punctuation
    feature set composed of various types of punctuation as derived from LIWC's
    punctuation categories. These punctuation categories include punctuation
    characters such as question marks, exclamation marks, periods, commas, dashes, etc.
\end{description}
\section{Experiments and Results}
We conduct several experiments with different
combinations of pre-processing steps and feature sets. 
Regarding machine learning algorithms, we use 
Support Vector Machine (SVM) with linear kernel, Random Forest (RF),
Logistic Regression (LR), Naive Bayes (NB) and Multilayer Perceptron (MLP), all of which
are provided by scikit-learn \cite{pedregosa2011scikit}.

Every classifier is trained on the predefined training split of the dataset
and validated against the predefined validation split, with 
accuracy, precision, recall and F1 measure.
In order to tune the classifiers, we employ a grid search method as offered by scikit-learn. 
Within this method, we first predefine sets of pre-processing steps, features, and 
hyperparameters. Then the grid search is started, where elements of these sets are 
automatically combined into different classification pipelines, which are then trained and evaluated using
respective training and validation data. After the grid search is finished, the best performing 
combination is retrieved. Table \ref{tbl:performance} shows results of highest performing
models in each machine learning approach as found via grid search.
\begin{table}[h!]
    \centering
    \caption{Performance on validation data in terms of accuracy (Acc), weighted average precision (P),
    weighted average recall (R) and weighted average F1 score (F1) of highest performing
    models in each machine learning approach as found via grid search.}
    \label{tbl:performance}
    \begin{tabular}{ *{5}{@{\hskip8pt}c@{\hskip8pt}} }
    \toprule
    \textbf{Algorithm} & \textbf{Acc} & \textbf{P} & \textbf{R} & \textbf{F1} \\
    \midrule
    SVM & 95.70   & 95.71  & 95.70  & 95.70    \\
    \midrule
    LR & 95.42  & 95.43 & 95.42 & 95.42     \\ 
    \midrule
    RF & 90.79  & 90.98 & 90.79 & 90.80     \\ 
    \midrule
    NB & 93.32  & 93.33 & 93.32 & 93.31     \\ 
    \midrule
    MLP & 93.60  & 93.62 & 93.60 & 93.59    \\ 
    \bottomrule
    \end{tabular}
\end{table}
\begin{figure}
    \centering
    \includegraphics[width=.8\linewidth]{"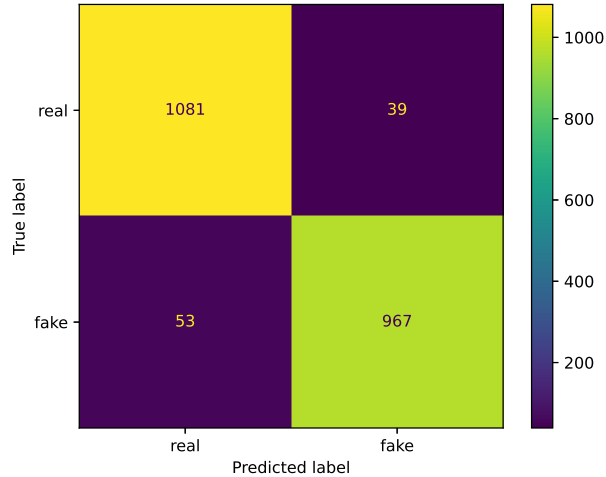"}
    \caption{Confusion matrix of best performing SVM model on validation data}
    \label{fig:svmcm}
\end{figure}
\\Using this method, we find the model with highest performance to be based on linear SVM. With a 
weighted average F1 score of 95.70\% on validation data, it is able to beat the 93.46\% 
baseline \cite{patwa2020fighting} that was set for this dataset. Our second best performing model 
is obtained via Logistic Regression (LR), followed
by Multilayer Perceptron (MLP), Naive Bayes (NB) and Random Forest (RF). Overall, only SVM and LR models were 
able to beat the baseline. The confusion matrix corresponding to SVM is illustrated in Figure \ref{fig:svmcm}.

In Table \ref{tbl:model}, we have a closer look at the concrete pre-processing and
feature combination that led to best performance in our SVM model and compare it to
other combinations. We observe that using  readability, punctuation and emotional features 
alone doesn't perform very well. In combination with each other, performance
increases noticeably, however, overall performance is still far away from top.
Interestingly, using n-gram features with pre-processing alone works
fairly well and already is enough to outperform the baseline. With the additional introduction
of the remaining features we reach our maximum performance.
\begin{table}[h!]
    \centering
    \caption{Different feature and pre-processing combinations as used in our linear SVM
    and corresponding model performance in terms of weighted average F1 score on validation data. 
    Model A describes the best feature and pre-processing combination as found by grid search.}
    \label{tbl:model}
    \begin{tabular}{ @{\hskip8pt}c@{\hskip8pt} p{9cm} *{1}{@{\hskip8pt}c@{\hskip8pt}} }
    \toprule
    \textbf{Model} & \textbf{Features and Pre-processing} & \textbf{F1} \\
    \midrule
    A &
    \textbf{N-grams:} Unigrams and bigrams
    are extracted from pre-processed
    social media posts. Pre-processing includes
    lowercase transformation, link removal, stop word removal,
    reply removal, XML entity replacement and lemmatization.
    For tokenization, NLTK's TweetTokenizer is used.
    \textbf{Readability:} Flesch Reading Ease is calculated
    on  pre-processed social media posts. Pre-processing includes
    lowercase transformation, link removal, reply removal and 
    XML entity replacement.
    \textbf{Punctuation and emotional tone:} Pre-processed social media posts
    are analyzed by LIWC and features are derived from LIWC's
    AllPunc, QMark and Exclam categories for punctuation and from 
    LIWC's Tone, affect, social and Authentic categories for emotional tone.
    Pre-processing includes lowercase transformation, link removal, reply removal and 
    XML entity replacement.
    & 95.70    \\
    \midrule
    B & Like A, but pre-processing consists solely of 
    lowercase transformation and XML entity replacement.
    &  94.66      \\ 
    \midrule
    C & Like A, but only readability features.  &  57.58      \\ 
    \midrule
    D & Like A, but only punctuation features.  &  49.17      \\ 
    \midrule
    E & Like A, but only emotional tone features.  &  59.22   \\ 
    \midrule
    F & C\texttt{+}D\texttt{+}E  &  67.21   \\
    \midrule
    H & Like A, but only n-grams features.  &  94.89   \\
    \bottomrule
    \end{tabular}
\end{table}
\subsection{Shared Task Submission}
In total, shared task participants were allowed to submit five system runs.
We decided to submit a system run for each of our five types of machine learning models.
To do that, we used the models from Table \ref{tbl:performance} and had them predict labels
for social media posts contained in the test set. Among our five submissions, the one using linear 
SVM yielded the best result with a weighted average F1 score of 95.19\%. This is consistent with
our expectations based on validation results, and is able to land a place in 
the middle of the leaderboard, where we are now placed at rank 80 of 167.
\section{Conclusion}
In  this  paper  we  presented  an  overview  of  our  participation 
at the Constraint 2021 COVID-19 Fake News Detection Shared Task.
We followed a classical machine learning approach, where we used Support Vector Machine,
Random Forest, Logistic Regression, Naive Bayes and Multilayer Perceptron
for classification. Feature-wise, we experimented with several linguistic features, 
such as n-grams, readability, emotional tone and punctuation. Our pre-processing pipeline
consisted of steps like stop word removal, stemming/lemmatization, link removal and more.
Each of the five classifiers we used was tuned by conducting a grid search over different combinations
of feature sets, pre-processing steps and hyperparameters. In the end, we submitted the best
run of each classifier and achieved the best result (95.19\% F1 score on test data) with our 
Support Vector Machine, landing on place 80 of 167 in the leaderboard.
\bibliographystyle{splncs03}
\bibliography{bibliography}

\begin{thebibliography}{10}
\providecommand{\url}[1]{\texttt{#1}}
\providecommand{\urlprefix}{URL }

\bibitem{barron2018overview}
Barr{\'o}n-Cede{\~n}o, A., Elsayed, T., Suwaileh, R., M{\`a}rquez, L.,
  Atanasova, P., Zaghouani, W., Kyuchukov, S., Da~San~Martino, G., Nakov, P.:
  Overview of the clef-2018 checkthat! lab on automatic identification and
  verification of political claims. task 2: Factuality. CLEF (Working Notes)
  2125 (2018)

\bibitem{coleman1975computer}
Coleman, M., Liau, T.L.: A computer readability formula designed for machine
  scoring. Journal of Applied Psychology  60(2),  283 (1975)

\bibitem{derczynski2017semeval}
Derczynski, L., Bontcheva, K., Liakata, M., Procter, R., Hoi, G.W.S., Zubiaga,
  A.: Semeval-2017 task 8: Rumoureval: Determining rumour veracity and support
  for rumours. arXiv preprint arXiv:1704.05972  (2017)

\bibitem{devlin2018bert}
Devlin, J., Chang, M.W., Lee, K., Toutanova, K.: Bert: Pre-training of deep
  bidirectional transformers for language understanding. arXiv preprint
  arXiv:1810.04805  (2018)

\bibitem{flesch2007flesch}
Flesch, R.: Flesch-kincaid readability test. Retrieved October  26(2007), ~3
  (2007)

\bibitem{flesch1948new}
Flesch, R.: A new readability yardstick. Journal of applied psychology  32(3),
  221 (1948)

\bibitem{ghanem2019upv}
Ghanem, B., Cignarella, A.T., Bosco, C., Rosso, P., Pardo, F.M.R.: Upv-28-unito
  at semeval-2019 task 7: Exploiting post’s nesting and syntax information
  for rumor stance classification. In: Proceedings of the 13th International
  Workshop on Semantic Evaluation. pp. 1125--1131 (2019)

\bibitem{gorrell2019semeval}
Gorrell, G., Kochkina, E., Liakata, M., Aker, A., Zubiaga, A., Bontcheva, K.,
  Derczynski, L.: Semeval-2019 task 7: Rumoureval, determining rumour veracity
  and support for rumours. In: Proceedings of the 13th International Workshop
  on Semantic Evaluation. pp. 845--854 (2019)

\bibitem{loper2002nltk}
Loper, E., Bird, S.: Nltk: the natural language toolkit. arXiv preprint
  cs/0205028  (2002)

\bibitem{marwick2017media}
Marwick, A., Lewis, R.: Media manipulation and disinformation online. New York:
  Data \& Society Research Institute  (2017)

\bibitem{doi:10.1177/0898264311428167}
Miller, L.M.S., Bell, R.A.: Online health information seeking: The influence of
  age, information trustworthiness, and search challenges. Journal of Aging and
  Health  24(3),  525--541 (2012),
  \url{https://doi.org/10.1177/0898264311428167}, pMID: 22187092

\bibitem{palen2008online}
Palen, L.: Online social media in crisis events. Educause quarterly  31(3),
  76--78 (2008)

\bibitem{patwa2021overview}
Patwa, P., Bhardwaj, M., Guptha, V., Kumari, G., Sharma, S., PYKL, S., Das, A.,
  Ekbal, A., Akhtar, S., Chakraborty, T.: Overview of constraint 2021 shared
  tasks: Detecting english covid-19 fake news and hindi hostile posts. In:
  Proceedings of the First Workshop on Combating Online Hostile Posts in
  Regional Languages during Emergency Situation ({CONSTRAINT}). Springer (2021)

\bibitem{patwa2020fighting}
Patwa, P., Sharma, S., PYKL, S., Guptha, V., Kumari, G., Akhtar, M.S., Ekbal,
  A., Das, A., Chakraborty, T.: Fighting an infodemic: Covid-19 fake news
  dataset (2020)

\bibitem{pedregosa2011scikit}
Pedregosa, F., Varoquaux, G., Gramfort, A., Michel, V., Thirion, B., Grisel,
  O., Blondel, M., Prettenhofer, P., Weiss, R., Dubourg, V., et~al.:
  Scikit-learn: Machine learning in python. the Journal of machine Learning
  research  12,  2825--2830 (2011)

\bibitem{pennebaker2015development}
Pennebaker, J.W., Boyd, R.L., Jordan, K., Blackburn, K.: The development and
  psychometric properties of liwc2015. Tech. rep. (2015)

\bibitem{pomerleau2017fake}
Pomerleau, D., Rao, D.: The fake news challenge: Exploring how artificial
  intelligence technologies could be leveraged to combat fake news. Fake News
  Challenge  (2017)

\bibitem{porter2002english}
Porter, M.F., Boulton, R., Macfarlane, A.: The english (porter2) stemming
  algorithm. Retrieved  18,  2011 (2002)

\bibitem{rangel2020overview}
Rangel, F., Giachanou, A., Ghanem, B., Rosso, P.: Overview of the 8th author
  profiling task at pan 2020: Profiling fake news spreaders on twitter

\bibitem{senter1967automated}
Senter, R., Smith, E.A.: Automated readability index. Tech. rep., CINCINNATI
  UNIV OH (1967)

\bibitem{thorne2018fact}
Thorne, J., Vlachos, A., Cocarascu, O., Christodoulopoulos, C., Mittal, A.: The
  fact extraction and verification (fever) shared task. arXiv preprint
  arXiv:1811.10971  (2018)

\bibitem{thorne2019second}
Thorne, J., Vlachos, A., Cocarascu, O., Christodoulopoulos, C., Mittal, A.: The
  second fact extraction and verification (fever2. 0) shared task. EMNLP 2019
  p.~1 (2019)

\bibitem{umer2020fake}
Umer, M., Imtiaz, Z., Ullah, S., Mehmood, A., Choi, G.S., On, B.W.: Fake news
  stance detection using deep learning architecture (cnn-lstm). IEEE Access  8,
   156695--156706 (2020)

\bibitem{zhou2019gear}
Zhou, J., Han, X., Yang, C., Liu, Z., Wang, L., Li, C., Sun, M.: Gear:
  Graph-based evidence aggregating and reasoning for fact verification (2019)

\end{thebibliography}
\end{document}